\documentclass[runningheads]{llncs}
\usepackage[T1]{fontenc}
\usepackage{graphicx}
\usepackage{microtype}
\usepackage{graphicx}
\usepackage{subfigure}
\usepackage{booktabs} % for professional tables
\usepackage{amsfonts} 
\usepackage{amsmath}
\usepackage{amssymb}
\usepackage{algorithm}
\usepackage{algpseudocode}
\usepackage{multirow}
\usepackage{hyperref}
\hypersetup{
    colorlinks=true,
    linkcolor=blue,
    filecolor=magenta,      
    urlcolor=cyan,
    }

\begin{document}
\title{A Penalty Approach for Normalizing Feature Distributions to Build Confounder-Free Models}
\titlerunning{A Penalty Approach for Normalizing Feature Distributions}
% If the paper title is too long for the running head, you can set
% an abbreviated paper title here
%
\author{Anthony Vento\inst{1}
%\orcidID{0000-1111-2222-3333} 
\and 
Qingyu Zhao\inst{1}
\orcidID{0000-0002-6368-0889} 
\and Robert Paul\inst{2}
%\orcidID{0000-1111-2222-3333} 
\and 
Kilian M. Pohl\inst{1,3}
\orcidID{0000-0001-5416-5159} 
\and
Ehsan Adeli\inst{1}
\orcidID{0000-0002-0579-7763}
}
% %
% index{Vento, Anthony}
% index{Zhao, Qingyu}
% index{Paul, Robert}
% index{Pohl, Kilian M.}
% index{Adeli, Ehsan}
\authorrunning{A. Vento et al.}
% % First names are abbreviated in the running head.
% % If there are more than two authors, 'et al.' is used.
% %
\institute{Stanford University, Stanford CA 94305, USA \and
 Missouri Institute of Mental Health, St. Louis MO 63121, USA \and SRI International, Menlo Park CA 94025, USA\\
 \email{eadeli@stanford.edu}
 }
\maketitle              % typeset the header of the contribution
\begin{abstract}
Translating machine learning algorithms into clinical applications requires addressing challenges related to interpretability, such as accounting for the effect of confounding variables (or metadata). Confounding variables affect the relationship between input training data and target outputs. When we train a model on such data, confounding variables will bias the distribution of the learned features. A recent promising solution, MetaData Normalization (MDN), estimates the linear relationship between the metadata and each feature based on a non-trainable closed-form solution. However,  this estimation is confined by the sample size of a mini-batch and thereby may cause the approach to be unstable during training. In this paper, we  extend the \underline{MDN} method by applying a \underline{P}enalty approach (referred to as \underline{PDMN}). We cast the problem into a bi-level nested optimization problem. We then approximate this optimization problem using a penalty method so that the linear parameters within the MDN layer are trainable and learned on all samples. This enables PMDN to be plugged into any architectures, even those unfit to run batch-level operations, such as transformers and recurrent models. We show improvement in model accuracy and greater independence from confounders using PMDN over MDN in a synthetic experiment and a multi-label, multi-site dataset of magnetic resonance images (MRIs). 

\end{abstract}
\section{Introduction}
% I

Modern machine learning approaches rely on automatically learning features from data \cite{zhong2016overview} using approaches such as convolutional neural networks (CNNs) \cite{adeli2020deep,deshmukh2022faster} and attention-based transformer models \cite{chen2021transunet,dosovitskiy2021an}. Although these methods solve challenging problems, they are known to capture spurious associations and biases introduced by confounding or protected variables  \cite{zhao2020training}. These limitations confine the neuroscientific impact of these algorithms, in which controlling for (and explaining the effects of) confounding variables is crucial. To remedy this, several approaches have been proposed, such as based on adversarial training \cite{zhao2020training,liu2018bridging}, counterfactual generative models \cite{neto2020causality,lahiri2022combining}, disentanglement \cite{tartaglione2021end,liu2021projection}, and correlation fair inference \cite{baharlouei2019r}. They learn features that are invariant or conditionally independent to the confounding variables. 

These training methods reduce the error from confounders with minimum compromise to model accuracy. However, adversarial models or those based on disentanglement and correlation are inefficient when accounting for multiple confounders (or metadata) and only partially remove the effects from feature maps of a single layer in the network \cite{zhao2020training}. Methods based on counterfactual require reliable generative models with respect to arbitrary variables, which is added complexity. To remove confounding effects at different feature layers, MetaData Normalization (MDN) \cite{lu2021metadata} can be plugged into a CNN and remove the effects of multiple confounders (or metadata) from the features while training the network. MDN aims to fix the distribution shift \cite{agarwal2021theory} caused by the confounding variables using a closed-form solution to linear regression capturing the relationship between confounders and each feature. 

The closed-form solution in MDN requires building a linear model (relationship between metadata and each feature) as a batch-level operation. It requires large batch sizes to obtain accurate approximations of the linear model. However, batch-level statistics in MDN (similar to batch normalization) face several challenges, including (1) instability when using small batch sizes, (2) increased training time due to the calculation of closed-form solutions for each feature at each iteration, (3) inconsistent results from training and inference since there are no batches during inference, (4) inability to use MDN for online training, in which the model is trained incrementally by feeding the samples in a sequential manner, and finally (5) inability to apply MDN to Recurrent Neural Networks \cite{ba2016layer} and selected transformer models \cite{vaswani2017attention,yao2021leveraging}. To overcome these limitations,  we now introduce a new penalty method that turns MDN to a layer with parameters that can be optimized with other components of the network during training. 

%To expand the application of MDN in modern architectures, we introduce a new penalty method that modifies MDN into a layer with trainable parameters that can be optimized with other components of the network during training. 
Referred to as a \underline{P}enalty approach for \underline{M}eta\underline{D}ata \underline{N}ormalizing (PMDN), our method improves upon the batch-level MDN operation. Specifically, PMDN can be applied to  all architectures and  any number of confounding variables. We show that PMDN is not dependent on the batch size. We apply PDMN to  a synthetic dataset to  analyze and validate the method within a controlled setting. We then examine  applicability of PDMN compared to MDN in  classifying multi-site MRIs into 4 diagnostic groups with image acquisition site, sex, and age as confounders.

\section{Methodology}
% put this at end of next paragraph 
% We first review the problem setup (Section 2.1)  and MDN (Section 2.2). Section 2.3 reformulates MDN into trainable parameters (Figure~\ref{fig:overview}).

\begin{figure}[h]
    \centering
    \includegraphics[width=0.99\linewidth]{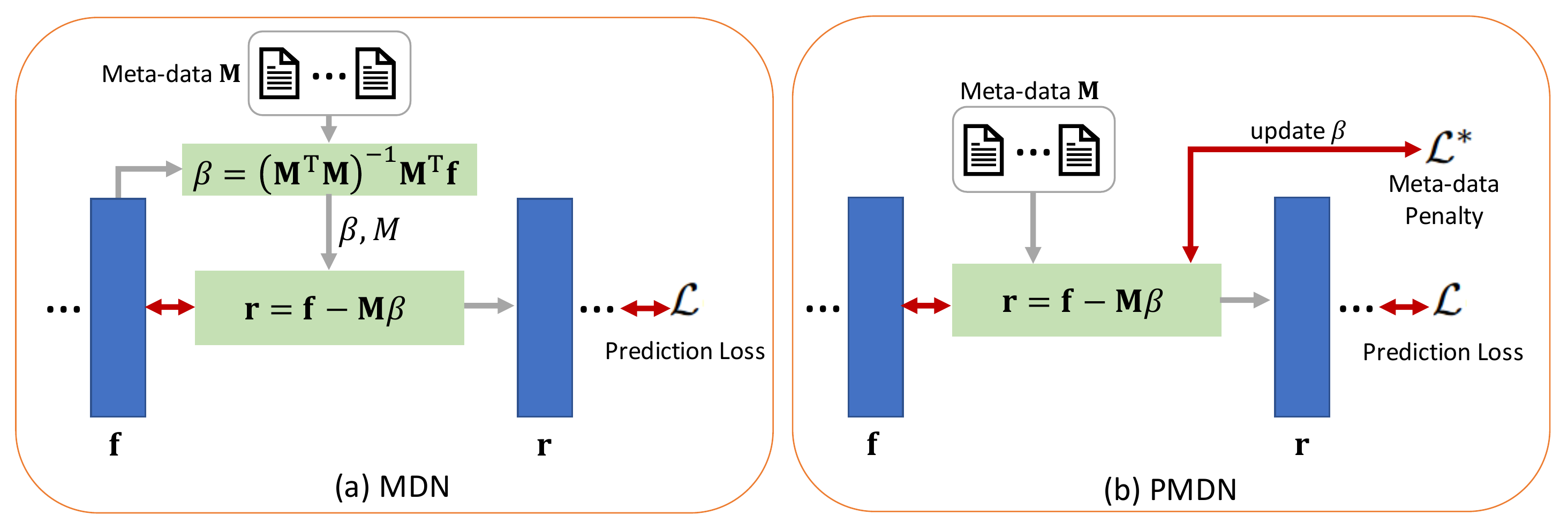}
    \caption{(a) MDN calculates residuals $\mathbf{r}$ based on the parameter $\beta$  being determined in closed-form, while PMDN (b) turns $\beta$ into network trainable parameters and simultaneously optimizes both features $f$ and $\beta$ by penalizing the prediction loss. }
    \label{fig:overview}
\end{figure}

% \subsection{Problem Setup}\label{MDNsetup}
% \vspace{5pt}\noindent\textbf{2.1 Problem Setup.}
Given a dataset of $N$ training samples, we define the metadata matrix as $\textbf{M} = [m_1, m_2, \dots, m_N]^{\top} \in \mathbb{R}^{N\times K}$. Each row of $\textbf{M}$, $m_i \in \mathbb{R}^K$, defines the metadata for sample $i$. Also, let $\textbf{f} = [f_1, f_2, \dots, f_N]^{\top} \in \mathbb{R}^N$ be the features for all training samples extracted at a particular layer. The goal of MDN is to remove confounding information from the features and use the residual component, $\textbf{r}$, as input to the next layer of the network.  The next subsection reviews how MDN performs this task via batch-level operations (Section 2.1), while Section 2.2 reformulates MDN so that it can be parameterized with respect to all training samples (Figure~\ref{fig:overview}). 

\subsection{MDN Review}\label{MDNreview}
% \vspace{5pt}\noindent\textbf{2.2 MDN Review.}
Lu et al. \cite{lu2021metadata} implemented the MDN layer as a general linear model (GLM), i.e., % if you have space 
\begin{equation}
\textbf{f} = \textbf{M} \beta + \textbf{r},    
\end{equation}
where $\beta \in \mathbb{R}^K$ is an unknown set of parameters, $\textbf{M} \beta$ describes the component in $\textbf{f}$ that is relevant to the metadata, and $\textbf{r}$ is the residual component that is irrelevant to the metadata. Then, the MDN operation is defined as 
\begin{equation}
\textbf{r} = \text{MDN}(\textbf{f} ;  \textbf{M} ) = \textbf{f}  -  \textbf{M} \beta.
\end{equation}
The MDN layer is not trained but instead $\beta$ is determined by the closed-form solution of least squares, i.e., 
\begin{equation}
    \beta_{ls} = (\textbf{M}^\top \textbf{M})^{-1}\textbf{M}^\top \textbf{f}.
\end{equation}
The underlying model assumes that the computation is performed across the features of  all $N$ training samples. However, training of deep learning is generally confined to batches of $b$ samples producing $\textbf{f} \in \mathbb{R}^{b}$. Therefore, \cite{lu2021metadata} approximates $\beta_{ls}$ as 
\begin{equation}
    \beta_{ls} = N (\textbf{M}^\top \textbf{M})^{-1} \mathbb{E}[m^\top f] \approx  \frac{N}{b}(\textbf{M}^\top \textbf{M})^{-1}\sum_{i=1}^b m_i f_i. 
    \label{eq:mdn}
\end{equation}
As Eq. \eqref{eq:mdn} approximates the expectation $\mathbb{E}[\cdot]$ only using data from a batch, the $\beta_{ls}$ estimates are generally inaccurate for a small batch size and are likely to vary from batch to batch resulting in model instability, similar to Batch Norm \cite{yong2020momentum}.

\subsection{PMDN: MDN as a Bi-Level Optimization}\label{MDN++}
% \vspace{5pt}\noindent\textbf{2.3 PMDN: MDN as a Bi-Level Optimization.}
To improve model stability and accuracy, we realize that inserting an MDN layer into a generic neural network is equivalent to reformulating the original objective function of the network to a bi-level optimization. 

Specifically, let $\textbf{X} = [x_1, x_2, \dots, x_N]^{\top}$ be the $N$ training samples and $\textbf{y} = [y_1,y_2,\dots,y_N]^{\top}$ be their corresponding prediction targets. Without loss of generality, let us assume that a network can be defined as the composition $\psi(\phi(\textbf{X}))$ between the first few layers $\phi$ and the layers afterwards $\psi$. For simplicity, we assume $\phi$ results in a one-channel feature but the following formulation generalizes to multi-channel features. Let $\textbf{W}$ be the network parameters of $\psi$ and $\phi$, then training of the network often reduces to solving the minimization problem 
\begin{equation}
\min_{\textbf{W}}\mathcal{L}(\psi(\phi(\textbf{X})),\textbf{y}).    
\end{equation}
Adding an MDN layer after $\phi$ changes  the minimization problem to
\begin{align}
\min_{\textbf{W}} & \mbox{ }\mathcal{L}(\psi(\phi(\textbf{X})-\textbf{M}\beta_{ls}),\textbf{y})\\
s.t. & \mbox{ }\beta_{ls} = \arg\min_{\beta} \mathcal{L}^*(\phi(\textbf{X});\textbf{M})=\arg\min_{\beta}||\phi(\textbf{X})-\textbf{M}\beta||^2. 
\label{eq:nested_constraint}
\end{align}
In other words, the constraint itself is a nested optimization, which aims to maximally remove the metadata effect from the feature learned by $\phi$.

To solve this bi-level optimization problem, PMDN (a Penalty approach for MDN) determines the minimum to a proxy objective function that combines the two minimization problems:
\begin{equation}
\min_{\beta, \textbf{W}} \mathcal{L}(\psi(\phi(\textbf{X})-\textbf{M}\beta),\textbf{y}) + \lambda \mathcal{L}^*(\phi(\textbf{X});\textbf{M}).
\label{eq:mdn_new}
\end{equation}
Now, Eq. \eqref{eq:mdn_new} is a well-defined, differentiable function that can be optimized by any gradient descent algorithm. Unlike MDN that sets $\beta$ to different values according to the batch construction, the $\beta$ estimates in PMDN can converge to a local optimum defined with respect to all training data.
% We observe that optimizing $\mathcal{L}^*$ simultaneously with respect to $\beta$ and $\textbf{W}$ would result in a trivial solution that simply shrinks the magnitude of $\phi(\textbf{X})$$ and $\beta$, so 
Here, we use an alternating optimization schema for removing metadata effects (Alg.~\ref{alg:pseudo}). As can be seen in lines 6 and 9, each of the two objectives have their own learning rates which are then consolidated into the optimizer (e.g., Adam \cite{kingma2014adam}), making the implementation independent from the hyperparameter $\lambda$. 
\begin{algorithm}[h]
\caption{Optimizing a network with PMDN}\label{alg:pseudo}
\begin{algorithmic}[1]
\Procedure{PMDN}{}
\State \textbf{Initialize:} network parameters $\textbf{W}$, PMDN parameters $\beta$, learning rates $\eta_1,\eta_2$  
\State \textbf{for} t \textbf{in} (0,1,$\cdots$,T):
\State $\quad$ Freeze $\textbf{W}^{(t)}$, Unfreeze $\beta^{(t)}$
\State $\quad$ $\hat{\textbf{y}} = \psi(\phi(\textbf{X})-\textbf{M}\beta^{(t)})$ \Comment{Forward pass}
\State $\quad$ $\beta^{(t+1)} = \beta^{(t)} - \eta_1\nabla_{\beta^{(t)}}\mathcal{L}^*(\phi(\textbf{X});\textbf{M})$
\State $\quad$ Freeze $\beta^{(t+1)}$, Unfreeze $\textbf{W}^{(t)}$
\State $\quad$ $\hat{\textbf{y}} = \psi(\phi(\textbf{X})-\textbf{M}\beta^{(t+1)})$ \Comment{Forward pass}
\State $\quad$ $\textbf{W}^{(t+1)} = \textbf{W}^{(t)} - \eta_2\nabla_{\textbf{W}^{(t)}}\mathcal{L}(\hat{\textbf{y}},\textbf{y})$
\State \textbf{end for}
\EndProcedure
\end{algorithmic}
\end{algorithm}
%\vspace{-5mm}
 
Although Alg.~\ref{alg:pseudo} only is based on one PMDN layer, multiple PMDN layers can be added without loss of generality to further remove any remaining residual confounding effects.  If we perform the metadata normalization after each of the $C$ features (from different layers or channels), $\mathcal{L}^*$ in Eq. \eqref{eq:mdn_new} is  the sum of all PMDN losses $\lambda \frac{1}{C}\sum_{i=1}^C||\textbf{f}^i - \textbf{M} \beta^i||^2$, where $\textbf{f}^i$ and $\beta^i$ are the feature vector and parameters of the $i^{\text{th}}$ PMDN, respectively. Furthermore, Alg.~\ref{alg:pseudo} uses Stochastic Gradient Descent (SGD) \cite{robbins1951stochastic} to update $\textbf{W}$ and $\beta$.  However, SGD can be replaced with other optimizers such as Adam \cite{kingma2014adam}. 

\section{Experiments}
We apply the method to  a synthetic and an MRI dataset with both continuous and discrete metadata.  For each experiment, we investigate the effect of metadata on a variety of architectures including a baseline CNN, the baseline network with MDN as described in Section 2.2, and the baseline  network with PMDN as described in Section 2.3. The code is available at \url{https://github.com/vento99/PMDN}.

\subsection{Synthetic Dataset Experiments}
\paragraph{Data.} The synthetic dataset \cite{lu2021metadata} consisted of 2000 $32\times32$ images subdivided into two groups of 1,000 images. The first group consisted of images where quadrants two and four are Gaussians with a variance sampled from the uniform distribution $\mathcal{U}(1,4)$. The second group consisted of images where quadrants two and four are Gaussians with a variance from $\mathcal{U}(3,6)$.  We introduce metadata into the third quadrant of the images.  In the first group, quadrant three also consists of a Gaussian with a variance from $\mathcal{U}(1,4)$ while in the second group, quadrant three consists of a Gaussian with a variance from $\mathcal{U}(3,6)$.  Theoretically, complete removal of the metadata effect will lead to a maximum model accuracy of $83.33\%$.

\paragraph{Implementation.} The baseline is a simple CNN of two standard blocks. The first block consists of two convolution layers with 16 and 32 filters and a ReLU activation after each convolution layer. The second block incorporates a fully connected layer of size 84 with ReLU activation followed by another fully connected layer. We use binary cross entropy loss as $\mathcal{L}$.  For all other methods, we add a normalization layer (one of BatchNorm \cite{ioffe2015batch}, MDN, or PMDN) after the convolution and first fully connected layers (before ReLU activations). 

Note that we also insert a LayerNorm layer \cite{ba2016layer} before each PMDN operation in order to stabilize the input features and to enable smoother gradients, faster training, and better generalization accuracy. Similar to the setup in \cite{lu2021metadata}, the metadata variable is colinear with the group labels.  Thus, the labels were included as an additional column in the metadata matrix $\textbf{M}$ during training to remove the metadata effect while preserving group differences. During inference, we remove the label column from $\textbf{M}$ and the last component from $\beta$ as implemented in \cite{lu2021metadata}.

\begin{figure}[t!]
    \centering
    \includegraphics[width=0.99\linewidth]{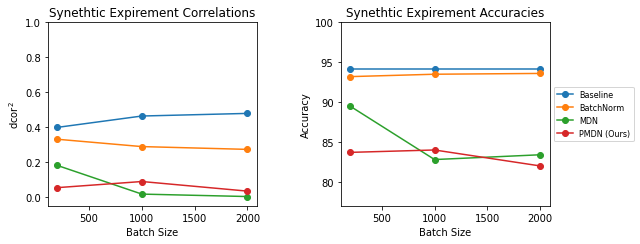}
    \caption{Effect of batch size on different normalization strategies. Left: dcor$^2$ values for different batch sizes.  Right: accuracy for different batch sizes.  Theoretically, the maximum accuracy  should be 83.3\% and higher values reflect prediction error resulting from the influence of metadata.}
    \label{fig:synthetic}
    % \vspace{-1mm}
\end{figure}

\paragraph{Evaluation.} To examine whether the metadata is removed from the learned features, we calculate the squared distance correlation (dcor$^2$) between the output of the first FC layer and the metadata of each group separately and report the average of the two dcor$^2$ values. Unlike linear correlation, dcor$^2$ examines relationships between high-dimensional variables.  Lower dcor$^2$ values reflect independence from  metadata confounding and thus, better normalization of  the feature distribution due to PMDN.

Figure~\ref{fig:synthetic} summarizes the results.  Results for the baseline, BatchNorm, and standard MDN layers are adopted from \cite{lu2021metadata}.  We see that the baseline and BatchNorm have an accuracy much greater than 83.3\% (the theoretical optimum), which means that the metadata effect has not been removed from the features. Instead, the model may have learned spurious associations between metadata and labels as the dcor$^2$ values are much higher than those for MDN and PMDN. 

For MDN, we see an inconsistency of results across different batch sizes. When the batch size is large, the batch-level closed-form solution of $\beta$ approximates the true $\beta$, so that MDN successfully normalizes the metadata effect. However, for a batch size of 200, MDN performance is significantly reduced. On the other hand, PMDN shows consistent results across all batch sizes supporting our hypothesis that the penalty-based approach is impartial to batch size.

\subsection{Multi-label Multi-site MRI Dataset Experiments}
\paragraph{Data.} The dataset consists of 1,262 T1-Weighted MRIs from three brain studies, where each MRI was bias field corrected, skull stripped, affinely registrated to a template of $64\times64\times64$ resolution. The three studies (see Table~\ref{tab:mristats} for summary) were performed by (1) the Memory and Aging Center, University of California - San Francisco (UCSF) \cite{zhang2016extracting}, (2) the Neuroscience Program, SRI International \cite{adeli2018chained}, and (3) the public Alzheimer’s Disease Neuroimaging Initiative (ADNI1) \cite{petersen2010alzheimer}. The participants of the three studies were divided into four cohorts: healthy older adults (no neurological/psychiatric diagnosis)  (CTRL; $N = 460$), adults infected with human immunodeficiency virus (HIV) without cognitive impairment (HIV; $N = 112$), HIV-infected individuals with cognitive impairment that were diagnosed with HIV-Associated Neurocognitive Disorder (HAND; $N = 145$) and HIV-negative adults diagnosed  with mild cognitive impairment (MCI) but no HIV (MCI; $N = 545$). MCI is a heterogeneous condition that reflects impairment in memory and other cognitive abilities \cite{delano2009heterogeneity}.  
%Historically, MCI was described as the prodromal stage of Alzheimer's disease (AD), characterized by impairment in memory and other cognitive abilities that was not of sufficient severity (at the time) to interfere with activities of daily living (e.g., driving, grooming, financial management). More recent studies recognize MCI as a heterogeneous condition that reflects AD neuropathology for a subset of individuals \cite{delano2009heterogeneity}.  
By definition, individuals with HAND meet the criteria for both MCI and HIV. Thus, this problem is formulated as a two-label classification problem: predicting whether or not individuals are infected with HIV and predicting whether or not individuals are diagnosed with cognitive impairment. 
% MRI data was harmonized from multiple sites to optimize the overall \textit{sample size}. 
For this dataset, the metadata includes the acquisition site (one-hot encoded), participant age ($z$-score) and participant self-identified sex (male/female). 

\begin{table}[t]
    \centering
    \setlength{\tabcolsep}{5.5pt}
    \caption{3D MRI Dataset Statistics}
    \label{tab:mristats} %\vspace{-8pt}
    \begin{tabular}{c|cccccc}
        \hline
        \textbf{Site} & \textbf{CTRL} & \textbf{MCI} & \textbf{HIV} & \textbf{HAND} &\textbf{F/M} & \textbf{Age Mean $\pm$ Std} \\
        \hline \hline
        UCSF & 156 & 148 & 37 & 145 &  97/389 & 67.00 $\pm$ 6.47\\
        ADNI & 229 & 397 & - & - & 253/373 & 75.16 $\pm$ 6.61\\
        SRI & 75 & - & 75 & - & 44/106 & 50.72 $\pm$ 11.33\\
        \hline
    \end{tabular}
    % \vspace{-9mm}
\end{table}

\paragraph{Implementation and Baseline Models.} The baseline consists of a 3D-ResNet \cite{3dresnet} followed by a series of fully connected (FC) blocks. The 3D-ResNet consists of four standard residual blocks. Each block incorporates a 3D Conv with ReLU activation and a skip connection. The number of filters for the standard convolutions in each block are 3, 6, 9, and 6, respectively.  All use kernel size 3 and padding size 1.  The output of the 3D-ResNet (flattened size 2048) is passed through a FC-ReLU-FC-ReLU-FC architecture. The FC outputs are of size 128, 16, and 2, respectively. The loss ($\mathcal{L}$) we use is the focal loss \cite{lin2017focal} to combat the class imbalance. For MDN, the layers are added after each FC-ReLU and after the final FC layer. As before with the Synthetic Dataset, for PMDN, we add a LayerNorm before the first two PMDN layers and include the labels as metadata. We also examine a BatchNorm architecture where we insert BatchNorms (BNs) after each FC-ReLU.  Finally, since most of the previous work focused on domain-adversarial methods for learning confounder-invariant features, we examine an adversarial training method similar to \cite{zhao2020training}. After the 3D-ResNet, we add an additional head of a FC-ReLU-FC, which attempts to predict the confounding variables. The correlation loss \cite{zhao2020training} from this head is adversarially subtracted from the classification loss when updating the weights of the 3D-ResNet.

\paragraph{Evaluation.} We perform 5-fold cross validation and report the results in Table~\ref{tab:hand}. We note that N=240 is the largest batch size feasible with our resource constraints. Based on the type of metadata variables (continuous, binary, or categorical), we choose different metric to investigate the metadata effect in the features. For age, we take the magnitude of the Pearson's correlation between the ages and each of the two output logits and report the average. For sex, we report on the  average magnitude of the point-biserial correlation between the sexes and each of the two output logits.  For site, we compute the average dcor$^2$ correlation between the site (one-hot encoded) and each of the two output logits. Finally, we calculate the accuracy for each of the four groups separately and report the average. 

For each batch size, PMDN achieves the highest accuracy. This highlights that PMDN mitigates the confounding effect and produces a less biased distribution. This is also evident in the low dcor$^2$ values for PMDN. As expected, small batch size significantly compromises MDN model performance. Additionally, we see that the adversarial training method \cite{zhao2020training} removed the confounding effects for sex and site but the confounding effect of age remained as noted by the correlations. This observation underscores the inherent limitation of the adversarial method in controlling multiple  confounds because each metadata variable  requires a new adversarial component in the network. Training multiple adversarial components sacrificed model accuracy.

\begin{table}[t]
    \centering
    \setlength{\tabcolsep}{5.4pt}
    \caption{Correlations with metadata variables and accuracies of all comparison methods for the 3D Multi-label MRI dataset across different batch sizes.}
    \label{tab:hand} %\vspace{-8pt}
    \begin{tabular}{c|r|ccccc}
        \hline
        \multirow{2}{*}{\textbf{|Batch|}} & \multirow{2}{*}{\textbf{Metric}} & \multirow{2}{*}{\textbf{Baseline}} &
        \multirow{2}{*}{\textbf{BN}} &
        \multirow{2}{*}{\textbf{Adversarial}} &
        \multirow{2}{*}{\textbf{MDN}} & \textbf{PMDN}\\
        & & & & & & \textbf{(Ours)}\\
        \hline
        \hline
        \multirow{ 4}{*}{20}& Age Corr $\downarrow$ & 0.431 & 0.382 & 0.408& 0.235 &\textbf{0.213}\\
        & Sex Corr $\downarrow$ & 0.209 & 0.237 & 0.154 & 0.172 &\textbf{0.141}\\
        & Site Corr $\downarrow$ & 0.388 & 0.312 &\textbf{0.086}&0.132& 0.155\\
        & Accuracy $\uparrow$ & 48.8\% & 44.0\% & 26.5\% & 41.2\% & \textbf{51.3}\%\\
        \hline
        \multirow{ 4}{*}{80}& Age Corr $\downarrow$ & 0.461 & 0.374 & 0.385& 0.225 &\textbf{0.208}\\
        & Sex Corr $\downarrow$ & 0.259 & 0.220 & 0.214 & 0.189 &\textbf{0.187}\\
        & Site Corr $\downarrow$ & 0.402 & 0.285 & 0.127 &\textbf{0.126}& 0.172\\
        & Accuracy $\uparrow$ & 49.7\% & 42.1\% & 25.8\% & 41.2\% & \textbf{50.7}\%\\
        \hline
        \multirow{ 4}{*}{160}& Age Corr $\downarrow$ & 0.488 & 0.384 & 0.543 &  0.241 &\textbf{0.199}\\
        & Sex Corr $\downarrow$ & 0.199 & 0.268 & \textbf{0.094} & 0.174& 0.188\\
        & Site Corr $\downarrow$ & 0.431 & 0.293 & 0.166 & 0.160 & \textbf{0.149}\\
        & Accuracy $\uparrow$ & 45.0\% &45.1\% & 26.8\% & 45.6\% &\textbf{ 50.7}\%\\
        \hline
        \multirow{ 4}{*}{240}& Age Corr $\downarrow$ & 0.488 & 0.369 & 0.382 & 0.185 &\textbf{0.180}\\
        & Sex Corr $\downarrow$ & 0.226 & 0.225 & 0.172 & 0.173 &\textbf{0.166}\\
        & Site Corr $\downarrow$ & 0.456 & 0.275 &\textbf{0.110}&0.137& 0.158\\
        & Accuracy $\uparrow$ & 43.5\% & 42.2\% & 27.8\% & 46.2\% & \textbf{51.9}\%\\
        \hline
    \end{tabular}
\end{table}

\begin{figure}[t]
    \centering
    \includegraphics[width=0.99\linewidth]{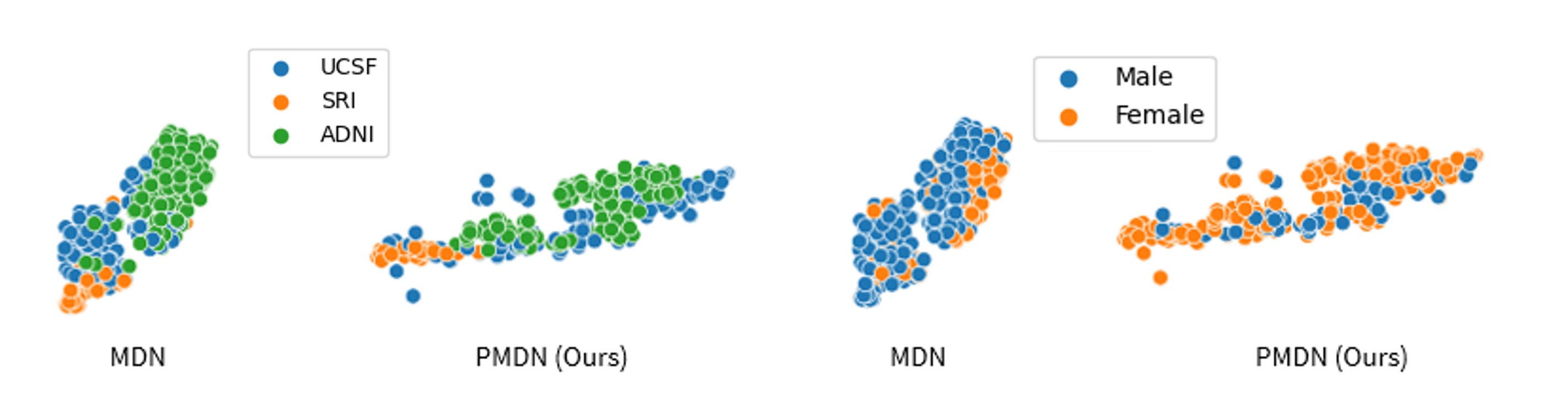}
    %\vspace{-15pt}
    \caption{tSNE visualization of features for MDN and PMDN (Left: site, Right: sex). }
    \label{fig:tsne}
    %\vspace{-5pt}
\end{figure}

Figure~\ref{fig:tsne} visualizes  the feature space (via tSNE \cite{van2008visualizing}) after removing metadata effects by MDN and PMDN for the small batch size of 80. As can be seen, the embedding space does not show a clear pattern with respect to sex differences (i.e., it is independent from the sex variable). For the case of site variable, PMDN illustrates less clustering effect compared to MDN's embedding. However, note that the three sites (UCSF, ADNI, SRI) are considerably different with respect to their class label distributions (see Table \ref{tab:mristats}), which explains moderate clustering in the embedding space. %In both cases, we see that the samples defined clearer clusters based on the confounding variables site and sex for the embedding generated by MDN compared to PMDN. 
In conclusion, this analysis qualitatively illustrates that of the two methods, PMDN is better at removing the confounding factor.

\section{Conclusion}
Herein, we introduce PMDN, our novel penalty method for removing bias in model training due to confounding factors. PMDN can be plugged into any neural network architecture and is independent from batch size. By removing the effects from confounding relationships between training and target outputs, PMDN minimizes the bias in the learned features. We show improvement of PMDN, a layer with trainable parameters, when compared to MDN, a layer with a closed-form solution, on a synthetic and a neuroimaging dataset.  The improvement in accuracy and confounder independence  from PMDN represent an important step towards neuroscience, imaging, or clinical applications of machine learning prediction models. 

\subsubsection{Acknowledgements} This study was partially supported by NIH Grants (AA017347, MH113406, and MH098759)
and Stanford Institute for Human-Centered AI (HAI) Google Cloud Platform (GCP) Credit.

%
% ---- Bibliography ----
%
% BibTeX users should specify bibliography style 'splncs04'.
% References will then be sorted and formatted in the correct style.
%
\bibliographystyle{splncs04}
\bibliography{paper1881}
\end{document}